\documentclass[runningheads]{llncs}
\usepackage{graphicx}
\usepackage{bm}
\usepackage{float} 
\usepackage{xcolor}
\usepackage{comment}
\usepackage{overpic}
\usepackage{amsmath}
\usepackage{amssymb}
\usepackage{bbding}
\usepackage{mathrsfs}
\usepackage{amsfonts}
\usepackage{graphicx}
\usepackage{appendix}
\usepackage{booktabs}
\usepackage{multirow}
\usepackage{url}
\usepackage{subfigure}
\usepackage{breakcites}
\usepackage[T1]{fontenc}
\usepackage[mathscr]{eucal}
\usepackage[font=footnotesize,labelfont=bf]{caption}
\usepackage{pifont}% http://ctan.org/pkg/pifont

\usepackage{geometry}
\begin{document}
\title{Transforming the Interactive Segmentation 
\\ for Medical Imaging}
\author{Wentao Liu\inst{1} \and
Chaofan Ma\inst{1} \and
Yuhuan Yang\inst{1} \and
Weidi Xie\inst{1,2} \and
Ya Zhang$^\dagger$\inst{1,2}}

\authorrunning{W. Liu et al. Transforming the Interactive Segmentation for Medical Imaging}

\institute{Cooperative Medianet Innovation Center, Shanghai Jiao Tong University %\\
%\email{ya\_zhang@sjtu.edu.cn}\\
 \and
Shanghai AI Laboratory
\\
\tt\small wtliu7@sjtu.edu.cn}


%
\maketitle              % typeset the header of the contribution

\def\thefootnote{}\footnotetext{$\dagger$ Corresponding author.}
\def\thefootnote{\arabic{footnote}}
\begin{abstract}
The goal of this paper is to interactively refine the automatic segmentation on challenging structures that fall behind human performance, 
either due to the scarcity of available annotations or the difficulty nature of the problem itself,
for example, on segmenting cancer or small organs.
Specifically, we propose a novel \textbf{T}ransformer-based architecture for \textbf{I}nteractive \textbf{S}egmentation~(\textbf{TIS}),
that treats the refinement task as a procedure for grouping pixels with similar features to those clicks given by the end users. 
Our proposed architecture is composed of Transformer Decoder variants, 
which naturally fulfills feature comparison with the attention mechanisms.
In contrast to existing approaches, 
our proposed TIS is not limited to binary segmentations,
and allows the user to edit masks for arbitrary number of categories.
To validate the proposed approach,
we conduct extensive experiments on three challenging datasets and demonstrate superior performance over the existing state-of-the-art methods.
 The project page is: \url{https://wtliu7.github.io/tis/}. 
\end{abstract}
\section{Introduction}
In medical image analysis, 
segmentation is undoubtedly one of the most widely researched problems in the literature.
The goal is to identify the structure of interest~(SOI) with pixel level accuracy, 
acquiring rich information, such as the position, size, and texture statistics, 
to assist clinicians for making assessments on diseases and better treatment plans.

In the recent years, 
tremendous progress has been made for {\em fully-automatic} segmentations 
by training deep neural networks on large-scale datasets in a {\em supervised manner},
for example, FCN~\cite{Long15}, UNet~\cite{Ronneberger15}, nnU-Net~\cite{Isensee20}.
Occasionally, 
such heavily-supervised approaches have already approached similar performance level as human expert.
However, apart from the well-solved problems, one critical issue remains, 
{\bf what else can we do to improve models' usability on the challenging scenarios},
where automatic predictions significantly under-perform humans, 
either due to the lack of large-scale training set, 
or the difficulty nature of problem, for example, on cancer segmentation.

One potential solution is interactive segmentation, 
with the goal of refining the automatic predictions by only a few user inputs, 
{\em e.g.}~clicks, scribbles, boundary delineation, etc.
In the literature, such line of research involves a long list of seminal works,
including the early attempts that took inspiration from mathematics and topology,
adopting variational methods~\cite{Chan01,Mumford89} to group pixels that share certain common features as the users' initialisation,
{\em e.g.}~intensity, texture, contrast, etc. 
However, such variational approaches usually involve heavy parameter-tuning 
and long inference time~({\em i.e.}~tens of minutes for large volumes), 
thus limiting its practical usefulness.
Till recently, approaches based on deep learning become popular, 
by transforming users' annotations into certain distance maps, 
such as euclidean~\cite{li2018interactive}, 
gaussian~\cite{maninis2018deep} or geodesic distance maps~\cite{wang2018deepigeos,luo2021mideepseg}, 
the ConvNets are trained to exploit such information together with images;
concurrently, a set of works have also considered to ease the interaction process,
for example, \cite{maninis2018deep,luo2021mideepseg} only requires to click on extreme points and~\cite{zhang2020interactive} use both positive and negative points.

Here, we continue this vein of research, 
and propose a novel {\bf T}ransformer-based architecture for {\bf I}nteractive {\bf S}egmentation, termed as {\bf TIS}.
In particular, for images that end up with unsatisfactory predictions, 
our proposed TIS only requires a few clicks from the end users, 
each pixel on the image only needs to compare with these ``examplar'' clicks, 
and copy the label information from its closest click.
Such ``compare-and-copy'' procedure can be elegantly achieved by adopting a variant of Transformer Decoder.
Notably, 
in contrast to the existing approaches that can only segment single structure at a time,
our TIS allows the end users to edit arbitrary number of categories simultaneously,
given each is provided with at least one examplar click.
We conduct evaluations on three challenging datasets and demonstrate superior performance over the state-of-the-art approaches.

\section{Methods}

In this paper, 
we consider a practical scenario for segmenting challenging structures in medical images, 
that is, to enable end users to correct the model's prediction with only {\em a few clicks} during inference time.

\subsection{Problem Scenario}

\begin{figure}[t]
\includegraphics[width=\textwidth]{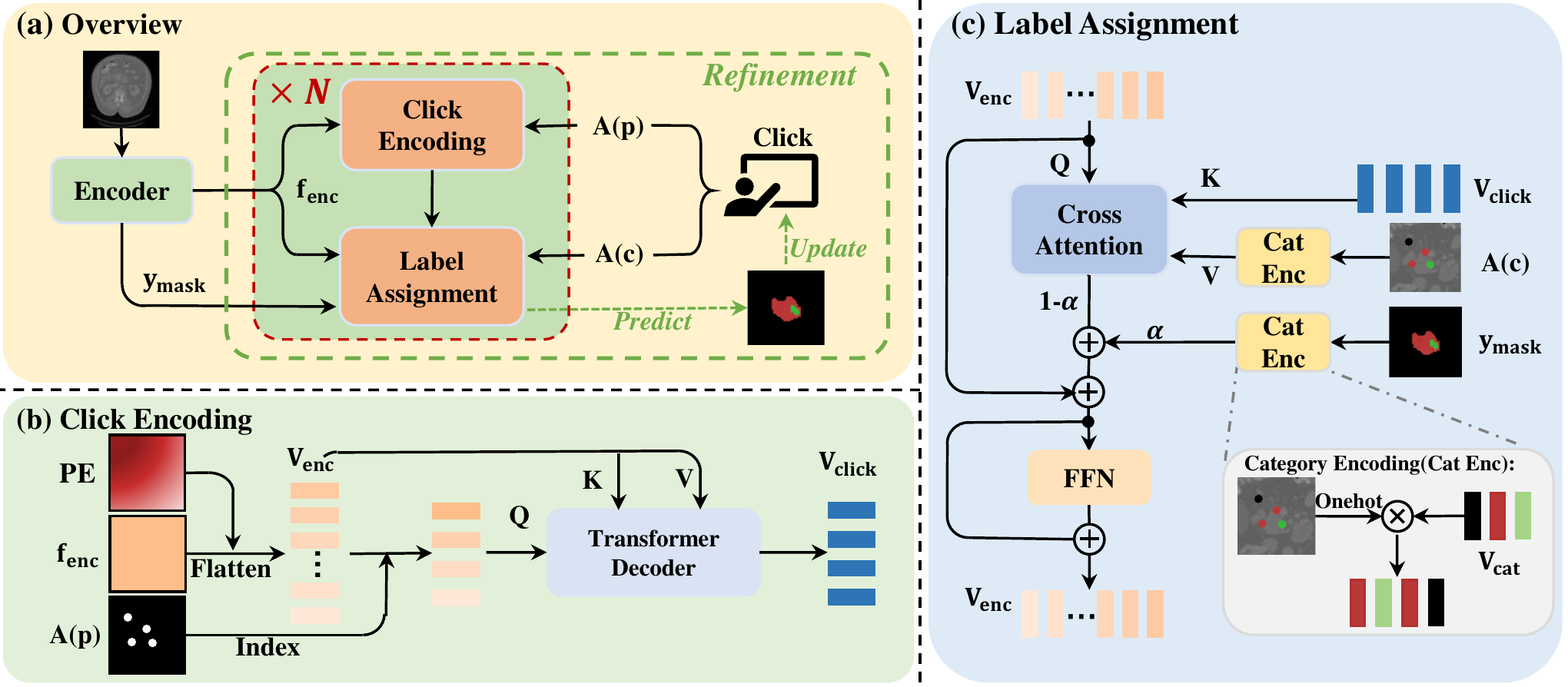}
\caption{Structure of our method. (a) Overview of the whole framework.  
(b) Structure of click encoding module. (d) Structure of label assignment module. 
} 
\label{overview}
\end{figure}

Given a \textbf{training set} of $n$ image-mask pairs, 
$\mathcal{D}_{\text{train}} = \{(\mathcal{I}_1, y_1), \allowbreak \dots, (\mathcal{I}_n, y_n)\}$, where $\mathcal{I} \in \mathbb{R}^{H \times W \times D }$ and 
$y \in \mathbb{R}^{H \times W \times D \times C}$, 
with $H, W, D, C$ denoting height, width, depth, and number of categories, respectively.
Our goal is to train a segmentation model that can not only give automatic predictions, 
but also allow the end users to refine the predictions during {\bf inference time}:
% ---------------------------------------------------------------------
\begin{align}
    \hat{y} = \mathrm{\Phi_{\text{REF}}}(\mathrm{\Phi_{\text{ENC}}}(\mathcal{I}; \Theta_e), \mathcal{A}; \Theta_r) \in \mathbb{R}^{H \times W \times D \times C},
\end{align}
% ---------------------------------------------------------------------
where $\hat{y}$ denotes the final prediction,
$\mathcal{A} = \{ (p_1, c_k), \dots, (p_n, c_k)\}$ refers the user's interaction in the form of pixel clicks,
with $p_i \in \mathbb{R}^{3}$ denoting the spatial position for each given click,
and $c_i \in \mathbb{R}^C$ referring the semantic category for the annotated pixel.
In the following sections, 
we will detail the building blocks of our proposed architecture~(as shown in Fig.~\ref{overview}(a)), 
namely, an encoder network~($\mathrm{\Phi}_{\text{ENC}}$, 
parameterized with $\Theta_e$) that extracts image features, 
and provides automatic structure segmentations;
and a refinement module~($\mathrm{\Phi}_{\text{REF}}$, parameterized with $\Theta_r$) that refines its prediction with the provided clicks from end users through {\em click encoding} and {\em label assignment}. 
Generally, the outputs from the encoder are architecture-agnostic, 
and experts can continuously interact with the refinement module until being satisfied.

% ------------------------------------------
\subsection{Encoder~($\mathrm{\Phi_{\text{ENC}}}$)} 
\label{sec:Encoder}
As for image encoder,
we adopt the popular nnU-Net~\cite{Isensee20}, 
that maps the input images to segmentation masks: 
\begin{align}
    \{y_{\text{mask}}, \text{\hspace{2pt}} f_{\text{enc}}\} = \mathrm{\Phi}_{\text{ENC}}(\mathcal{I};  \text{\hspace{2pt}}\Theta_e),
\end{align}
where $y_{\text{mask}} \in \mathbb{R}^{H \times W \times D \times C}$ refers to the output mask with $C$ categories,
and $f_{\text{enc}} \in \mathbb{R}^{\frac{H}{2} \times \frac{W}{2} \times \frac{D}{2} \times m}$ denotes the dense feature embeddings from the penultimate layer of nnU-Net. 
During training, the encoder is simply trained with standard pixelwise cross-entropy loss. \\

% ------------------------------------------
\subsection{Refinement~($\mathrm{\Phi_{\text{REF}}}$)}\label{sec:Decoder}
At a high-level,
our method is based on the intuition that users' click can be treated as a set of ``examplars'', 
and each pixel to be segmented  can simply compare with these ``examplars'',
and copy the category information from its closest clicks.
Technically, we adopt an architecture that interleaves Transformer Decoder variants, 
alternating between {\em click encoding} and {\em label assignment}, as detailed below.

% ------------------------------------------
\paragraph{\bf Tokenization: }
To start with, we vectorize the encoder's output encodings: 
\begin{align}
        \mathcal{V}_{\text{enc}} &= \phi_{\text{flatten}}(f_{\text{enc}}) + \phi_{\text{PE}}(\Omega),
\end{align}
where $\phi_{\text{flatten}}$ refers to a reshape operation, and $\phi_{\text{PE}}$ refers to the learnable positional encodings for a dense 3D grid~($\Omega$), 
$\mathcal{V}_{\text{enc}} \in \mathbb{R}^{HWD \times m}$ ends up to be the tokenized vector sequences of the encoder embeddings. 
% ------------------------------------------
\paragraph{\bf Click Encoding: }
To incorporate the users' feedback in refinement, we encode these clicks into vectors:
\begin{align}
        & \mathcal{V}_{\text{click}} = \phi_{\text{index}}(f_{\text{enc}}, \mathcal{A}(p)),\\
        & \mathcal{V}_{\text{click}} = \psi_{\text{T-Dec}}(\underbrace{W^q_1 \cdot \mathcal{V}_{\text{click}}}_{\text{Query}},  
        \text{\hspace{3pt}} \underbrace{W^k_1 \cdot \mathcal{V}_{\text{enc}}}_{\text{Key}}, 
        \text{\hspace{3pt}} \underbrace{W^v_1 \cdot \mathcal{V}_{\text{enc}}}_{\text{Value}}),
\end{align}
where $\phi_{\text{index}}$ refers to an indexing function that simply pick out the vectors from the dense feature map, based on corresponding spatial locations of the clicks.
To avoid notation abuse, 
we use $\mathcal{V}_{\text{click}} \in \mathbb{R}^{k \times m}$ to represent the results from indexing initial click embeddings, and the one after enrichment from a standard Transformer Decoder~($\psi_{\text{T-Dec}}$), 
with input {\em Query} derived from a linear projection of the click embeddings,
while {\em Key} and {\em Value} are generated by applying two different linear transformations on the encoder's outputs.

In specific, 
the Transformer Decoder consists of multi-head cross-attention (MHCA), multi-head self-attention (MHSA), 
a feed-forward network (FFN), and residual connections, 
which effectively enriches the click embeddings by aggregating information from the encoder outputs. 
For more details, we would refer the readers to the original Transformer paper~\cite{Vaswani17}.

% ------------------------------------------
\paragraph{\bf Label Assignment: }
Here, we construct a ``compare-and-copy'' mechanism, 
that assigns labels to each pixel based on two factors:
(1) the similarity between pixels and users' clicks,
(2) the weighting of the automatic segmentation. 

Specifically, we adopt a Transformer Decoder variant, as shown below:
\begin{gather*}
     \mathcal{V}_{\text{enc}} = \alpha \cdot \psi_{\text{T-Dec}}(\underbrace{W^q_2 \cdot \mathcal{V}_{\text{enc}}}_{\text{Query}},  
         \text{\hspace{3pt}} \underbrace{W^k_2 \cdot \mathcal{V}_{\text{click}}}_{\text{Key}}, 
         \text{\hspace{3pt}} 
         \underbrace{\phi_{\text{CE}}(\mathcal{A}(c))}_{\text{Value}}) +(1-\alpha) \cdot \phi_{\text{CE}}(y_{\text{mask}}),
\end{gather*}
where $\phi_{\text{CE}}(\cdot)$ refers to a projection~(accomplished by a learnable MLP) on the category labels to high-dimensional embeddings, operating on both user's click and automatic segmentations, these embeddings are then used for constructing the {\em Value} in Transformer Decoder. As for {\em Query} and {\em Key}, 
they are computed by applying linear transformations on the dense features and click embeddings respectively.
{\bf Note that}, with such architecture design,
the cross attention effectively computes a similarity matching between each pixel~($\mathcal{V}_{
\text{enc}}$) and users' click~($\mathcal{V}_{\text{click}}$), 
and copying back the ``category'' information, 
balanced with a {\em learnable} weighting scalar~($\alpha$)  between predictions and user's clicks.

\vspace{-3pt}
\paragraph{\bf Training: }
As mentioned above, features after label assignment now have incorporated the category information obtained from both automatic segmentation and clicks.
To properly train the model, we simulate the user interactions at training stage,
where clicks are sampled based on the discrepancy between automatic prediction and the groundtruth annotations, {\em i.e.}~erroneous predictions. 
After stacking 6 layers of {\em click encoding} and {\em label assignment} modules, we adopt a {\bf linear MLP layer} to predict the segmentation, 
and train it with pixelwise cross-entropy loss.

\vspace{-3pt}
\paragraph{\bf Discussion: }
Inspired by the observation that pixels of the same category should ideally be clustered together in some high-dimensional space, 
we thus adopt a variant of Transformer Decoder, which naturally facilitates the ``compare-and-copy'' mechanism, 
{\em i.e.}~compute similarity between pixels and user's click, 
then copy the category information correspondingly.
Note that, such procedure works in the same manner for segmentation of arbitrary class, this is in contrast to the existing approaches that are only limited to work for binary segmentation.

\section{Experiments}

\subsection{Datasets}

In this paper, we conduct experiments on 
the Medical Segmentation Decathlon (MSD) datasets~\cite{antonelli2021medical}.
Specifically, we focus on three challenging subsets:

\paragraph{\bf Lung (training set) } 
consists of preoperative thin-section CT scans from 63 patients with non-small cell lung cancer. 
The goal is to segment the tumors within the lung (L1).
We randomly split into 50 cases for training and the rest 13 cases for evaluation.

\vspace{-3pt}
\paragraph{\bf Colon (training set) } 
consists of 126 portal venous phase CT scans of patients undergoing resection of primary colon cancer.
The goal is to segment the colon cancer primaries (L1). 
100 cases are split for training randomly and the remaining 26 for evaluation.

\vspace{-3pt}
\paragraph{\bf Pancreas (training set) } 
consists of 281 portal venous phase CT scans of patients undergoing resection of pancreatic masses. The goal is to segment {\em both} pancreatic parenchyma (L1) and pancreatic tumor (L2).
224 cases are randomly picked for training and the remaining 57 for evaluation.

% -------------------------------------
\subsection{Evaluation Metrics}
For quantitative evaluation, 
we employ Dice Similarity Coefficient (DSC):
\begin{equation}
\operatorname{DSC}\left(\mathcal{R}_{p}, \mathcal{R}_{g}\right)=\frac{2\left|\mathcal{R}_{p} \cap \mathcal{R}_{g}\right|}{\left|\mathcal{R}_{p}\right|+\left|\mathcal{R}_{g}\right|},
\end{equation}
where $\mathcal{R}_{p}, \mathcal{R}_{g}$ represent the region of prediction and the ground-truth, respectively. $|\cdot|$ is the number of pixels/voxels in the corresponding region. 
And the goal is thus to get a higher accuracy with less user clicks.

\subsection{Implementation Details}
We use nnU-Net~\cite{Isensee20} as our encoder, and retrain it on corresponding datasets under the default settings. 
According to the official code of nnU-Net\footnote{\url{https://github.com/MIC-DKFZ/nnUNet.}}, the setting we used to train it is that the ``network'' is ``3d-fullres'', the ``network-trainer'' is ``nnUNetTrainerV2'' and leave other options as default. Due to the complexity of Transformer and the cost of memory, we used feature of the penultimate layer of nnU-Net. In practice, the feature will be cropped based on the automatic segmentation and clicks before processed by our model.

In practice, the feature will be cropped based on the automatic segmentation and clicks before processed by our model.
In the experiment, we train our model for 200 epochs. 
It is optimized using AdamW optimizer, 
starting with a learning rate of $10^{-3}$, and decreasing with a rate factor 0.9 every 10 epochs.

\begin{table}[t]
\centering
\setlength\tabcolsep{8pt}
\resizebox{.9\textwidth}{!}{%
\begin{tabular}{l|c|c|c|c|c|c}
\toprule
\multirow{2}{*}{Metric}& \multirow{2}{*}{Method}& \multirow{2}{*}{Year}& \multicolumn{1}{c|}{Lung}  & \multicolumn{1}{c|}{Colon} & \multicolumn{2}{c}{Pancreas} \\
\cmidrule{4-7}         
& & &\multicolumn{1}{c|}{L1} & \multicolumn{1}{c|}{L1} & \multicolumn{1}{c}{L1} & \multicolumn{1}{c}{L2} \\
\midrule
\multirow{6}{*}{Dice}& Automatic~\cite{Isensee20} & 2018 & 64.99$$ & 44.84$$ $$ & 82.16$$ & 49.34\\
& InterCNN~\cite{bredell2018iterative} & 2018 & 80.07$_{\pm2.65}$ & 69.58$_{\pm2.97}$ & 82.31$_{\pm3.28}$ & 74.17$_{\pm2.91}$ \\
& DeepIGeoS~\cite{wang2018deepigeos} & 2019& 81.74$_{\pm1.72}$ & 70.61$_{\pm2.46}$ & 82.77$_{\pm1.51}$ & 75.36$_{\pm2.60}$ \\
& BS-IRIS~\cite{ma2020boundary}   & 2020 & 81.67$_{\pm2.14}$ & 71.27$_{\pm1.82}$ & 85.16$_{\pm1.34}$ & 76.49$_{\pm2.48}$ \\
& MIDeepSeg~\cite{luo2021mideepseg} & 2021& 82.31$_{\pm3.58}$ & 71.89$_{\pm3.09}$ & 84.69$_{\pm4.03}$ & 70.34$_{\pm4.36}$   \\
& {\bf Ours}    & 2022 & {\bf 85.07}$_{\pm1.55}$ & {\bf 73.03}$_{\pm1.68}$ & {\bf 87.72}$_{\pm1.28}$ & {\bf 77.91}$_{\pm2.07}$   \\ 
\bottomrule
\end{tabular}}%
\vspace{5pt}
\caption{Performances of different methods on three datasets with 10 clicks. }
\label{tab:10clickspreformance}
\vspace{-0.7cm}
\end{table}

\section{Results}
\subsection{User Interactions Simulation}
Following previous works~\cite{wang2018deepigeos,ma2020boundary,luo2021mideepseg}, 
we also adopt a robust agent to simulate the user clicks.
The clicking positions are chosen as the center region of the largest mis-segmented regions according to the ground-truth.
For each clicking step, a small disturbance $\epsilon$ (10 pixels) is added on each click's positions to imitate the behavior of a real user and force the model to be more robust. 
Additionally, 
in cases when the centroid is not within the mis-segmented region, 
then we pick an arbitrary point in the non-boundary region as the click.

\subsection{Comparisons with State-of-the-Art}
\paragraph{\bf Performance on Different Datasets: }
As shown in Tabel~\ref{tab:10clickspreformance}, 
given 10 clicks, our approach achieves the best performance in all three challenging datasets. 
It is worth noting that the ``Automatic'' row shows the results from state-of-the-art automatic segmentation model~(nnU-Net~\cite{Isensee20}), 
while our proposed TIS can improve the prediction by a large margin with only a few clicks.

\begin{figure}[t]
\includegraphics[width=\textwidth]{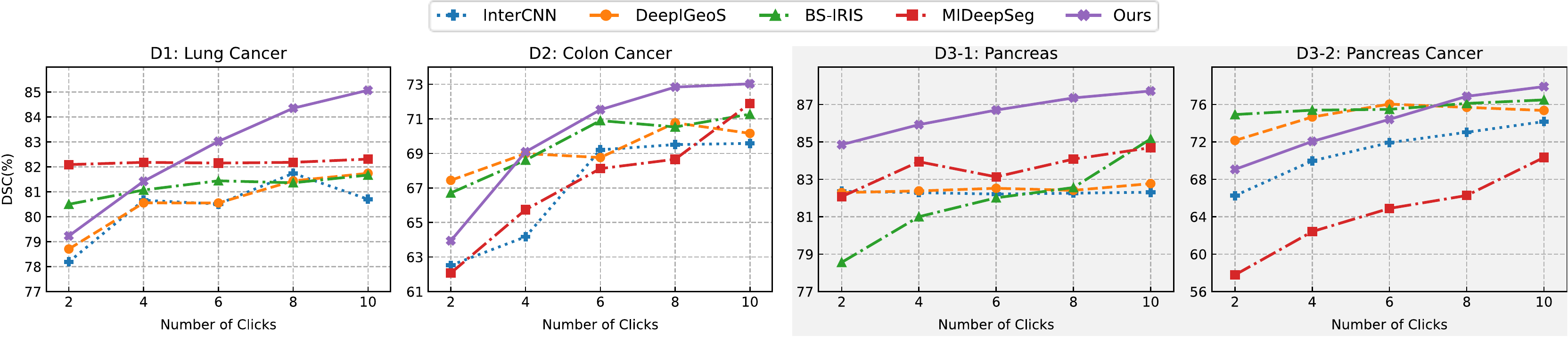}
\caption{Improvements in one interactive sequence of different methods on three datasets (four structures). Note that figures with gray background are two different structures in the same dataset.
This figure is best viewed on pdf.} 
\label{improvements_different}
\end{figure}

\vspace{-3pt}
\paragraph{\bf Improvements in One Interaction Sequence: }
In Fig.~\ref{improvements_different},
we plot the results with multi-round refinement for different approaches.
As can be seen, 
despite all interactive segmentation approaches have shown improvements with interaction, 
our proposed TIS maintains the growing trend with more clicks available, 
and substantially outperforms the others in the end. 
Specifically, for datasets that have small regions and heterogeneous appearance (for example, cancers), 
our method has smoother and more robust refinements through interaction.

\subsection{Ablation Study}
In this section, 
we perform a detailed ablation study by removing the component from TIS. 
All the ablation experiments are performed on the Pancreas dataset.

\vspace{-3pt}
\begin{table}[!htb]
\centering
\setlength\tabcolsep{5pt}
\resizebox{.8\textwidth}{!}{%
\begin{tabular}{cc|cc|cc}
\hline
\multicolumn{2}{c|}{Modules} & \multicolumn{2}{c|}{5 clicks} & \multicolumn{2}{c}{10 clicks} \\ \hline
click encoding       & label assignment      & pancreas       & cancer      & pancreas       & cancer      \\ \hline
\XSolidBrush& \Checkmark     &  82.74         & 66.74              &  83.32            &  71.52         \\
\Checkmark & \XSolidBrush    &   81.23        & 70.54              &  80.78             &  69.82         \\
\Checkmark & \Checkmark      & {\bf 86.15}       & {\bf  72.82}          & {\bf 87.72}           &{\bf  77.91 }      \\ \hline
\end{tabular}}
\vspace{5pt}
\caption{Quantitive ablation study on critical modules.}\label{ablation}
\vspace{-1.2cm}
\end{table}

\paragraph{\bf Effect of Click Encoding: } 
The main purpose of click encoding is to add context information to the click embeddings. 
To validate its effect, we only index embeddings, 
but do not encode them through Transformer Decoder~($\psi_{\text{T-Dec}}$).
As shown in Table.~\ref{ablation},
the performance has a severe recession,
due to the lack of context information, 
especially  after feature update of label assignment,
the network degrades, refining only a  small area around the click.

\paragraph{\bf Effect of Label Assignment: }
In this part, 
we {\bf do not} encode the category label as {\em Value} , 
rather, 
use a MLP projection, as normally did in standard transformer. 
Through a stack of Transformer layers, 
the cross-attention still computes the similarity between pixels and clicks, but unable to directly copy the label information.
And at the end,
the click embeddings obtained from click encoding are used as classifiers to segment based on similarity.
In order to make a fair comparison, we still use the weighting scalar~($\alpha$) to combine the automatic segmentation and the Transformer segmentation.  
As shown in Table.~\ref{ablation},
the performance has dropped drastically.
With the number of clicks increasing, it even gets worse.
In practice, despite we click at the mis-segmented region, the network is still unable to correct it.

\subsection{Visualization of Results}

\begin{figure}[t]
\includegraphics[width=.97\textwidth]{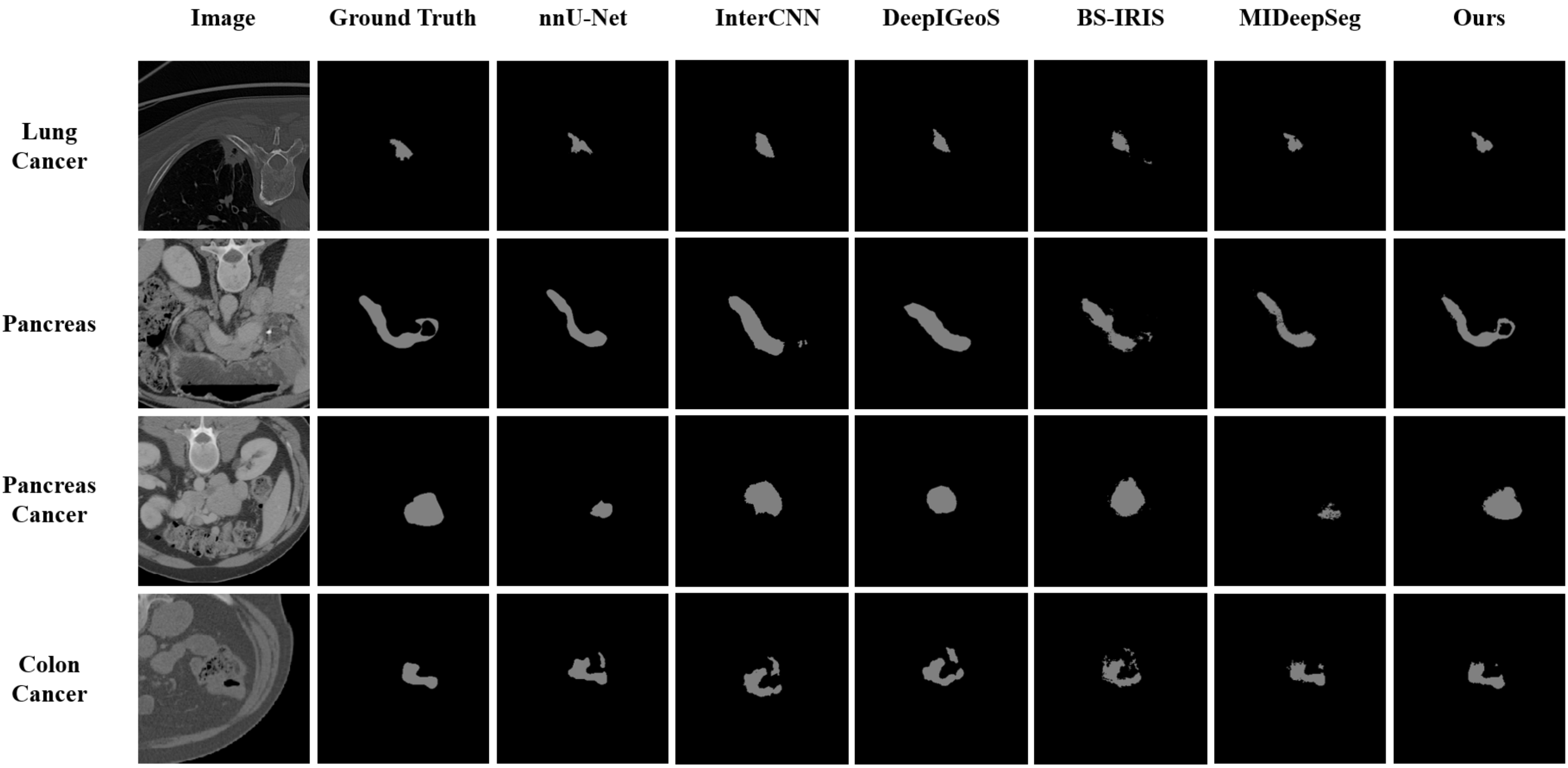}
\caption{Visualization of qualitative comparison on three datasets (four structures). All interactive methods are provided with 10 clicks.} 
\label{visualization}
\vspace{-2pt}
\end{figure}

Fig.~\ref{visualization} shows a comparison of the final segmentation results for the automatic nnU-net and four interactive methods. 
It can be seen that the state-of-the-art automatic segmentation method nnU-net often fails in these challenging scenarios, showing the potential of interactive segmentation methods.
And our method, with the same 10 clicks provided, have a more detailed segmentation and largest performance improvements compared with all interactive methods, especially for some small regions like lung cancer and colon cancer.

\subsection{Visualization of the Interaction Process}

\begin{figure}[h]
\centering
\includegraphics[width=0.93\textwidth]{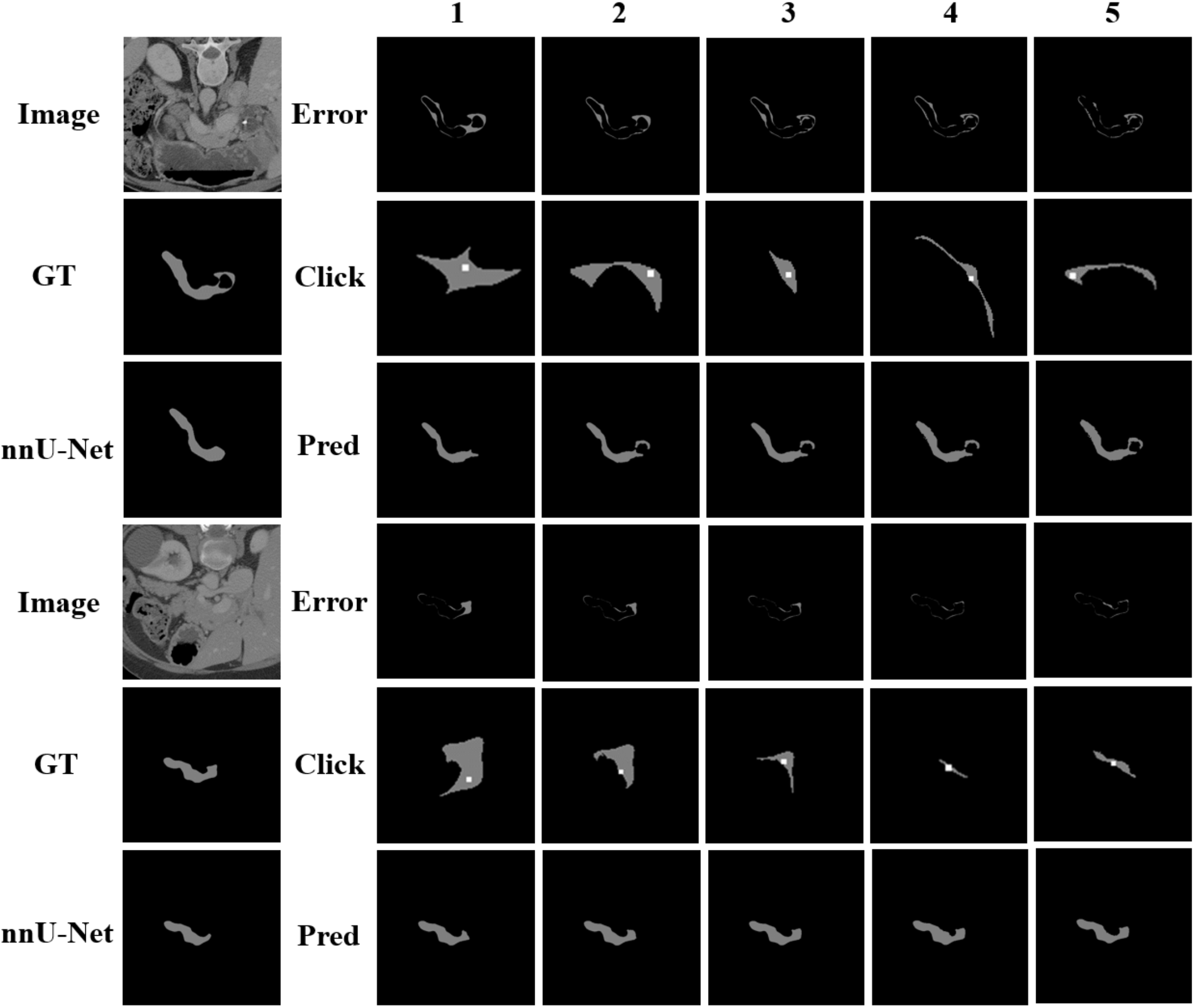}
\caption{Visualization of the interaction process with 5 clicks. Two examples are shown. } 
\label{appendix}
\end{figure}

Fig.~\ref{appendix} shows the interaction process, where two examples from Pancreas dataset are detailed. For each step, we provide the mis-segmented region~(``Error''), the click position~(``Click'') and the prediction of our method~(``Pred''). The performance improves as the number of clicks increasing, with the mis-segmented region decreases alongside.

\section{Conclusion}

We propose a novel Transformer-based framework~({\bf TIS}) for interactive segmentation.
It is designed for a realistic and challenging scenario, 
where automatic segmentations largely under-perform human performance.
The core idea is to treat the users' clicks as ``examplars'', 
and our proposed method is able to segment SOI by comparing each pixel on the image with the provided clicks, and copy the label information from them.
Specifically, our method interleaves two variants of Transformer Decoders, 
alternating between {\em click encoding} and {\em label assignment}.
We validate our method on three challenging datasets and demonstrate superior performance over the existing state-of-the-art methods.
Additionally, 
our methods breaks through the limitation that previous methods can only complete single classification.


\bibliographystyle{splncs04}
\bibliography{bib/ref}

\begin{thebibliography}{10}
\providecommand{\url}[1]{\texttt{#1}}
\providecommand{\urlprefix}{URL }
\providecommand{\doi}[1]{https://doi.org/#1}

\bibitem{antonelli2021medical}
Antonelli, M., Reinke, A., Bakas, S., Farahani, K., Landman, B.A., Litjens, G.,
  Menze, B., Ronneberger, O., Summers, R.M., van Ginneken, B., et~al.: The
  medical segmentation decathlon. arXiv preprint arXiv:2106.05735  (2021)

\bibitem{bredell2018iterative}
Bredell, G., Tanner, C., Konukoglu, E.: Iterative interaction training for
  segmentation editing networks. In: International workshop on machine learning
  in medical imaging. pp. 363--370. Springer (2018)

\bibitem{Chan01}
Chan, T.F., Vese, L.A.: Active contours without edges. IEEE Transactions on
  Image Processing  (2001)

\bibitem{Isensee20}
Isensee, F., Jaeger, P.F., Kohl, S.A.A., Petersen, J., Maier-Hein, K.H.:
  nnu-net: a self-configuring method for deep learning-based biomedical image
  segmentation. Nature Methods  (2020)

\bibitem{li2018interactive}
Li, Z., Chen, Q., Koltun, V.: Interactive image segmentation with latent
  diversity. In: Proceedings of the IEEE Conference on Computer Vision and
  Pattern Recognition. pp. 577--585 (2018)

\bibitem{Long15}
Long, J., Shelhamer, E., Darrell, T.: Fully convolutional models for semantic
  segmentation. In: Proceedings of the IEEE Conference on Computer Vision and
  Pattern Recognition (2015)

\bibitem{luo2021mideepseg}
Luo, X., Wang, G., Song, T., Zhang, J., Aertsen, M., Deprest, J., Ourselin, S.,
  Vercauteren, T., Zhang, S.: Mideepseg: Minimally interactive segmentation of
  unseen objects from medical images using deep learning. Medical Image
  Analysis  \textbf{72},  102102 (2021)

\bibitem{ma2020boundary}
Ma, C., Xu, Q., Wang, X., Jin, B., Zhang, X., Wang, Y., Zhang, Y.:
  Boundary-aware supervoxel-level iteratively refined interactive 3d image
  segmentation with multi-agent reinforcement learning. IEEE Transactions on
  medical imaging  \textbf{40}(10),  2563--2574 (2020)

\bibitem{maninis2018deep}
Maninis, K.K., Caelles, S., Pont-Tuset, J., Van~Gool, L.: Deep extreme cut:
  From extreme points to object segmentation. In: Proceedings of the IEEE
  Conference on Computer Vision and Pattern Recognition. pp. 616--625 (2018)

\bibitem{Mumford89}
Mumford, D., Shah, J.: Optimal approximations by piecewise smooth functions and
  associated variational problems. Communications on Pure and Applied
  Mathematics  (1989)

\bibitem{Ronneberger15}
Ronneberger, O., Fischer, P., Brox, T.: U-net: Convolutional networks for
  biomedical image segmentation. In: Proceedings of the International
  Conference on Medical Image Computing and Computer Assisted Intervention
  (2015)

\bibitem{Vaswani17}
Vaswani, A., Shazeer, N., Parmar, N., Uszkoreit, J., Jones, L., Gomez, A.N.,
  Kaiser, L., Polosukhin, I.: Attention is all you need. In: Advances in Neural
  Information Processing Systems (2017)

\bibitem{wang2018deepigeos}
Wang, G., Zuluaga, M.A., Li, W., Pratt, R., Patel, P.A., Aertsen, M., Doel, T.,
  David, A.L., Deprest, J., Ourselin, S., et~al.: Deepigeos: a deep interactive
  geodesic framework for medical image segmentation. IEEE transactions on
  pattern analysis and machine intelligence  \textbf{41}(7),  1559--1572 (2018)

\bibitem{zhang2020interactive}
Zhang, S., Liew, J.H., Wei, Y., Wei, S., Zhao, Y.: Interactive object
  segmentation with inside-outside guidance. In: Proceedings of the IEEE
  conference on computer vision and pattern recognition. pp. 12234--12244
  (2020)

\end{thebibliography}
% ---- Appendix ----
% \include{sec/05-appendix}
\end{document}